\documentclass[sigconf]{acmart}

\AtBeginDocument{%
  \providecommand\BibTeX{{%
    \normalfont B\kern-0.5em{\scshape i\kern-0.25em b}\kern-0.8em\TeX}}}

\copyrightyear{2020}
\acmYear{2020}
\setcopyright{acmcopyright}
\acmConference[MM '20] {28th ACM International Conference on Multimedia}{October 12--16, 2020}{Seattle, WA, USA.}
\acmBooktitle{28th ACM International Conference on Multimedia (MM '20), October 12--16, 2020, Seattle, WA, USA.}
\acmPrice{15.00}
\acmDOI{10.1145/3394171.3413965}
\acmISBN{978-1-4503-7988-5/20/10}

\begin{document}
	\fancyhead{}

\title{HiFaceGAN: Face Renovation via \\ Collaborative Suppression and Replenishment}

\author{Lingbo Yang}
\orcid{0000-0002-5339-925X}
\authornote{Equal contribution. Authors are ordered horizontally, just ignore the above format.}
\author{Shanshe Wang}
\author{Siwei Ma}
\authornote{Corresponding author.}
\author{Wen Gao}
\affiliation{%
	\institution{Institute of Digital Media, Peking University, Beijing, China}
	\postcode{100871}
}
\email{{lingbo,sswang,swma,wgao}@pku.edu.cn}

\author{Chang Liu}
\authornotemark[1]
\affiliation{%
	\institution{University of Chinese \\Academy of Sciences}
	\city{Beijing}
	\state{China}
	\postcode{101408}
}
\email{liuchang615@mails.ucas.ac.cn}

\author{Pan Wang}
\author{Peiran Ren}
\affiliation{%
	\institution{DAMO Academy, Alibaba Group}
	\city{Hangzhou}
	\state{China}
	\postcode{101408}
}
\email{{dixian.wp,peiran.rpr}@alibaba-inc.com}

\def\TODO[#1]{\textcolor{red}{#1}}
\def\NOTE[#1]{(\textcolor{blue}{#1})}

\def \etal{~\emph{et. al. }}
\def \L{\mathcal{L}}

\def \check {$\checkmark$}
\def \cross {$\times$}

\renewcommand{\shortauthors}{Yang and Liu, et al.}

\begin{abstract}
	
	Existing face restoration researches typically rely on either the image degradation prior or explicit guidance labels for training, which often lead to limited generalization ability over real-world images with heterogeneous degradation and rich background contents. In this paper, we investigate a more challenging and practical ``dual-blind'' version of the problem by lifting the requirements on both types of prior, termed as ``Face Renovation''(FR). Specifically, we formulate FR as a semantic-guided generation problem and tackle it with a collaborative suppression and replenishment (CSR) approach. This leads to HiFaceGAN, a multi-stage framework containing several nested CSR units that progressively replenish facial details based on the hierarchical semantic guidance extracted from the front-end content-adaptive suppression modules. Extensive experiments on both synthetic and real face images have verified the superior performance of our HiFaceGAN over a wide range of challenging restoration subtasks, demonstrating its versatility, robustness and generalization ability towards real-world face processing applications. Code is available at \href{https://github.com/Lotayou/Face-Renovation}{\underline{https://github.com/Lotayou/Face-Renovation}}.
	
\end{abstract}




\maketitle

\section{Introduction}

\begin{figure*}
	\includegraphics[width=\textwidth]{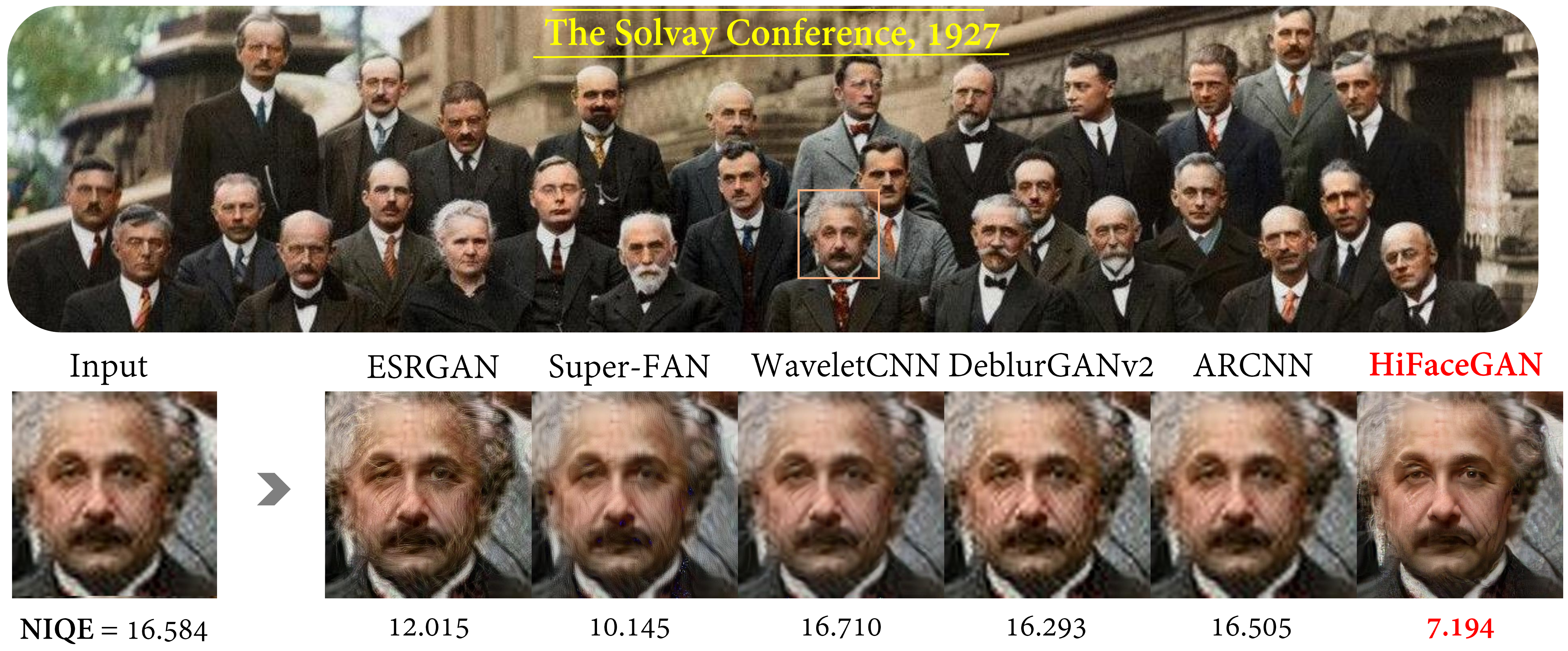}
	\caption{Face renovation results of related state-of-the-art methods. Our HiFaceGAN achieves the best perceptual quality as measured by the Naturalness Image Quality Evaluator(NIQE) ~\cite{niqe}. \emph{(Best to view on the computer screen for your convenience to zoom in and compare the quality of facial details. Ditto for other figures.)}
	}
	\label{fig:teaser}
\end{figure*}

Face photographs record long-lasting precious memories of individuals and historical moments of human civilization. Yet the limited conditions in the acquisition, storage, and transmission of images inevitably involve complex, heterogeneous degradations in real-world scenarios, including discrete sampling, additive noise, lossy compression, and beyond. With great application and research value, face restoration has been widely concerned by industry and academia, as a plethora of works ~\cite{ir_review}\cite{deblurring_survey}\cite{com_art_reduction_survey} devoted to address specific types of image degradation. Yet it still remains a challenge towards more generalized, unconstrained application scenarios, where few  works can report satisfactory restoration results.

For face restoration, most existing methods typically work in a ``non-blind'' fashion with specific degradation of prescribed type and intensity, leading to a variety of sub-tasks including super resolution\cite{sr_review}\cite{srcnn}\cite{edsr}\cite{esrgan}, hallucination~\cite{fh_review}\cite{face_hallu_4}, denoising\cite{RIDNet}\cite{VDNet}, deblurring~\cite{deblurring_survey}\cite{DeblurGAN}\cite{DeblurGANv2} and compression artifact removal~\cite{com_art_reduction_survey}\cite{ARCNN}\cite{EPGAN}. However, task-specific methods typically exhibit poor generalization over real-world images with complex and heterogeneous degradations. A case in point shown in Fig. ~\ref{fig:teaser} is a historic group photograph taken at the Solvay Conference, 1927, that super-resolution methods, ESRGAN~\cite{esrgan} and Super-FAN~\cite{SuperFAN}, tend to introduce additional artifacts, while other three task-specific restoration methods barely make any difference in suppressing degradation artifacts or replenishing fine details of hair textures, wrinkles, etc., revealing the impracticality of task-specific restoration methods.

When it comes to blind image restoration~\cite{BIR_original}, researchers aim to recover high-quality images from their degraded observation in a ``single-blind'' manner without \emph{a priori} knowledge about the type and intensity of the degradation. It is often challenging to reconstruct image contents from artifacts without degradation prior, necessitating additional guidance information such as categorial~\cite{CategorySpecificID} or structural prior~\cite{FSRNet} to facilitate the replenishment of faithful and photo-realistic details. For blind face restoration~\cite{MultiScaleFR-AAAI2018}\cite{SGEN}, facial landmarks~\cite{SuperFAN}, parsing maps~\cite{pix2pixHD}, and component heatmaps~\cite{faceSR_ECCV2018} are typically utilized as external guidance labels. In particular, Li \emph{et.al.} explored the guided face restoration problem~\cite{BlindFR-ECCV2018}\cite{EnhancedBlindFR-CVPR2020}, where an additional high-quality face is utilized to promote fine-grained detail replenishment. However, it often leads to limited feasibility for restoring photographs without ground truth annotations. Furthermore, for real-world images with complex background, introducing unnecessary guidance could lead to inconsistency between the quality of renovated faces and unattended background contents.

In this paper, we formally propose ``Face Renovation''(FR), an extra challenging, yet more practical task for photo-realistic face restoration under a ``dual-blind'' condition, lifting the requirements of both the degradation and structural prior for training. Specifically, we formulate FR as a semantic-guided face synthesis problem, and propose to tackle this problem with a collaborative suppression and replenishment(CSR) framework. To implement FR, we propose HiFaceGAN, a generative framework with several nested CSR units to perform face renovation in a multi-stage fashion with hierarchical semantic guidance. Each CSR unit contains a suppression module for extracting layered semantic features with content-adaptive convolution, which are utilized to guide the replenishment of corresponding semantic contents. Extensive experiments are conducted on both the synthetic FFHQ~\cite{stylegan_ffhq} dataset and real-world images against competitive degradation-specific baselines, highlighting the challenges in the proposed face renovation task and the superiority of our framework. In summary, our contributions are threefold:

\begin{itemize}
	\item We present a challenging yet practical task, termed as ``Face Renovation (FR)'', to tackle unconstrained face restoration problems in a ``dual-blind'' fashion, lifting the requirements on both degradation and structure prior.
	\item We propose the well-designed HiFaceGAN, a collaborative suppression and replenishment (CSR) framework nested with several CSR modules for photorealistic face renovation. 
	\item Extensive experiments are conducted on both synthetic and real face images with significant performance gain over a variety of ``non-blind'' and ``single-blind'' baselines, verifying the versatility, robustness and generalization capability of the proposed HiFaceGAN.
\end{itemize}

\section{Related Works} 

\subsection{Non-Blind Face Restoration}
Image restoration consists of a variety of subtasks, such as denoising~\cite{RIDNet}\cite{VDNet}, deblurring~\cite{DeblurGAN}\cite{DeblurGANv2} and compression artifact removal~\cite{ARCNN}\cite{EPGAN}.
In particular, image super resolution~\cite{srcnn}\cite{edsr}\cite{srgan}\cite{esrgan} and its counterpart for faces, hallucination~\cite{fh_review}\cite{face_hallu_1}\cite{face_hallu_3}\cite{face_hallu_4}, can be considered as specific types of restoration against downsampling.
However, existing works often works in a ``non-blind'' fashion by prescribing the degradation type and intensity during training, leading to dubious generalization ability over real images with complex, heterogeneous degradation. In this paper, we perform face renovation by replenishing facial details based on hierarchical semantic guidance that are more robust against mixed degradation, and achieves superior performance over a wide range of restoration subtasks against state-of-the-art ``non-blind'' baselines.

\subsection{Blind Face Restoration}
Blind image restoration ~\cite{BIR_original} \cite{BIR_1}\cite{BIR_2} aims to directly learn the restoration mapping based on observed samples. However, most existing methods for general natural images are still sensitive to the degradation profile~\cite{BlindIR-wo-prior} and exhibit poor generalization over unconstrained testing conditions.
For category-specific~\cite{CategorySpecificID} (face) restoration, it is commonly believed that incorporating external guidance on facial prior would boost the restoration performance, such as semantic prior~\cite{ComponentSP}, identity prior~\cite{face_hallu_1}, facial landmarks~\cite{SuperFAN}\cite{FSRNet} or component heatmaps~\cite{faceSR_ECCV2018}. In particular, Li~\emph{et.al.}~\cite{BlindFR-ECCV2018} explored the guided face restoration scenario with an additional high-quality guidance image to help with the generation of facial details. Other works utilize objectives related to subsequent vision tasks to guide the restoration, such as semantic segmentation~\cite{whenID_meets_high} and recognition~\cite{Joint_BIR_recog}. In this paper, we further explore the ``dual-blind'' case targeting at unconstrained face renovation in real-world applications. Particularly, we reveal an astonishing fact that with collaborative suppression and replenishment, the dual-blind face renovation network can even outperform state-of-the-art ``single-blind'' methods due to the increased capability for enhancing non-facial contents. This brings fresh new insights for tackling unconstrained face restoration problem from a generative view.

\subsection{Deep Generative Models for Face Images}

Deep generative models, especially GANs~\cite{gan} have greatly facilitated conditional image generation tasks~\cite{pix2pix}\cite{cyclegan}, especially for high-resolution faces~\cite{progressivegan}\cite{stylegan_ffhq}\cite{stylegan2}. Existing methods can be roughly summarized into two categories: semantic-guided methods utilize parsing maps~\cite{pix2pixHD}, edges~\cite{vid2vid}, facial landmarks~\cite{SuperFAN} or anatomical action units~\cite{GANimation} to control the layout and expression of generated faces, and style-guided generation~\cite{stylegan_ffhq}\cite{stylegan2} utilize adaptive instance normalization~\cite{adain} to inject style guidance information into generated images. Also, combining semantic and style guidance together leads to multi-modal image generation~\cite{bicyclegan}, enabling separable pose and appearance control of the output images. Inspired by SPADE~\cite{spade} and SEAN~\cite{sean} for semantic-guided image generation based on \emph{external} parsing maps, our HiFaceGAN utilizes the SPADE layers to implement collaborative suppression and replenishment for multi-stage face renovation, which progressively replenishes plausible details based on hierarchical semantic guidance, leading to an automated renovation pipeline without external guidance.

\section{Face Renovation}\label{sec:face_renovation}
Generally, the acquisition and storage of digitized images involves many sources of degradations, including but not limited to discrete sampling, camera noise and lossy compression. Non-blind face restoration methods typically focus on reversing a specific source of degradation, such as super resolution, denoising and compression artifact removal, leading to limited generalization capability over varying degradation types, Fig~\ref{fig:teaser}. On the other hand, blind face restoration often relies on the structural prior or external guidance labels for training, leading to quality inconsistency between foreground and background contents.
To resolve the issues in existing face restoration works, we present face renovation to explore the capability of generative models for ``dual-blind'' face restoration without degradation prior and external guidance.
Although it would be ideal to collect authentic low-quality and high-quality image pairs of real persons for better degradation modeling, the associated legal issues concerning privacy and portraiture rights are often hard to circumvent.
In this work, we perturb a challenging, yet purely artificial face dataset~\cite{stylegan_ffhq} with heterogeneous degradation in varying types and intensities to simulate the real-world scenes for FR. Thereafter, the methodology and comprehensive evaluation metrics for FR are analyzed in detail.

\subsection{Degradation Simulation}\label{sec:train_setting}

With richer facial details, more complex background contents, and higher diversity in gender, age, and ethnic groups, the synthetic dataset FFHQ~\cite{stylegan_ffhq} is chosen for evaluating FR models with sufficient challenges. We simulate the real-world image degradation by perturbing the FFHQ dataset with different types of degradations corresponding to respective face processing subtasks, which will be also evaluated upon our proposed framework to demonstrate its versatility. For FR, we superimpose four types of degradation (except 16x mosaic) over clean images in random order with uniformly sampled intensity to replicate the challenge expected for real-world application scenarios.
\footnote{The python script will be provided in supplementary materials.}
Fig.~\ref{fig:deg_egs} displays the visual impact of each type of degradation upon a clean input face. It is evident that mosaic is the most challenging due to the severe corruption of facial boundaries and fine-grained details. Blurring and down-sampling are slightly milder, with the structural integrity of the face almost intact. Finally, JPEG compression and additive noise are the least conceptually obtrusive, where even the smallest details (such as hair bang) are clearly discernable. As will be evidenced later in Sec.~\ref{sec:sota}, the visual impact is consistent with the performance of the proposed face renovation model. Finally, the full degradation for FR is more complex and challenging than all subtasks (except 16x mosaic), with both additive noises/artifacts and detail loss/corruption. We believe the proposed degradation simulation can provide sufficient yet still manageable challenge towards real-world FR applications.

\subsection{Methodology}
With the single dominated type of degradation, existing methods are devoted to fit an inverse transformation to recover the image content. When it comes to real-world scenes, the low-quality facial images usually contain unidentified heterogeneous degradation, necessitating a unified solution that can simultaneously address common degradations without prior knowledge.
Given a severely degraded facial image, the renovation can be reasonably decomposed into two steps, 1) suppressing the impact of degradations and extracting robust semantic features; 2) replenishing fine details in a multi-stage fashion based on extracted semantic guidance.
Generally speaking, a facial image can be decomposed into semantic hierarchies, such as structures, textures, and colors, which can be captured within different receptive fields~\cite{Marr1982VisionAC}. Also, noise and artifacts need to be adaptively identified and suppressed according to different scale information.
This motivates the design of HiFaceGAN, a multi-stage renovation framework consisting of several nested collaborative suppression and replenishment(CSR) units that is capable of resolving all types of degradation in a unified manner. Implementation details will be introduced in the following section.

\begin{figure}[!pt]
	\centering
	\includegraphics[width=\linewidth]{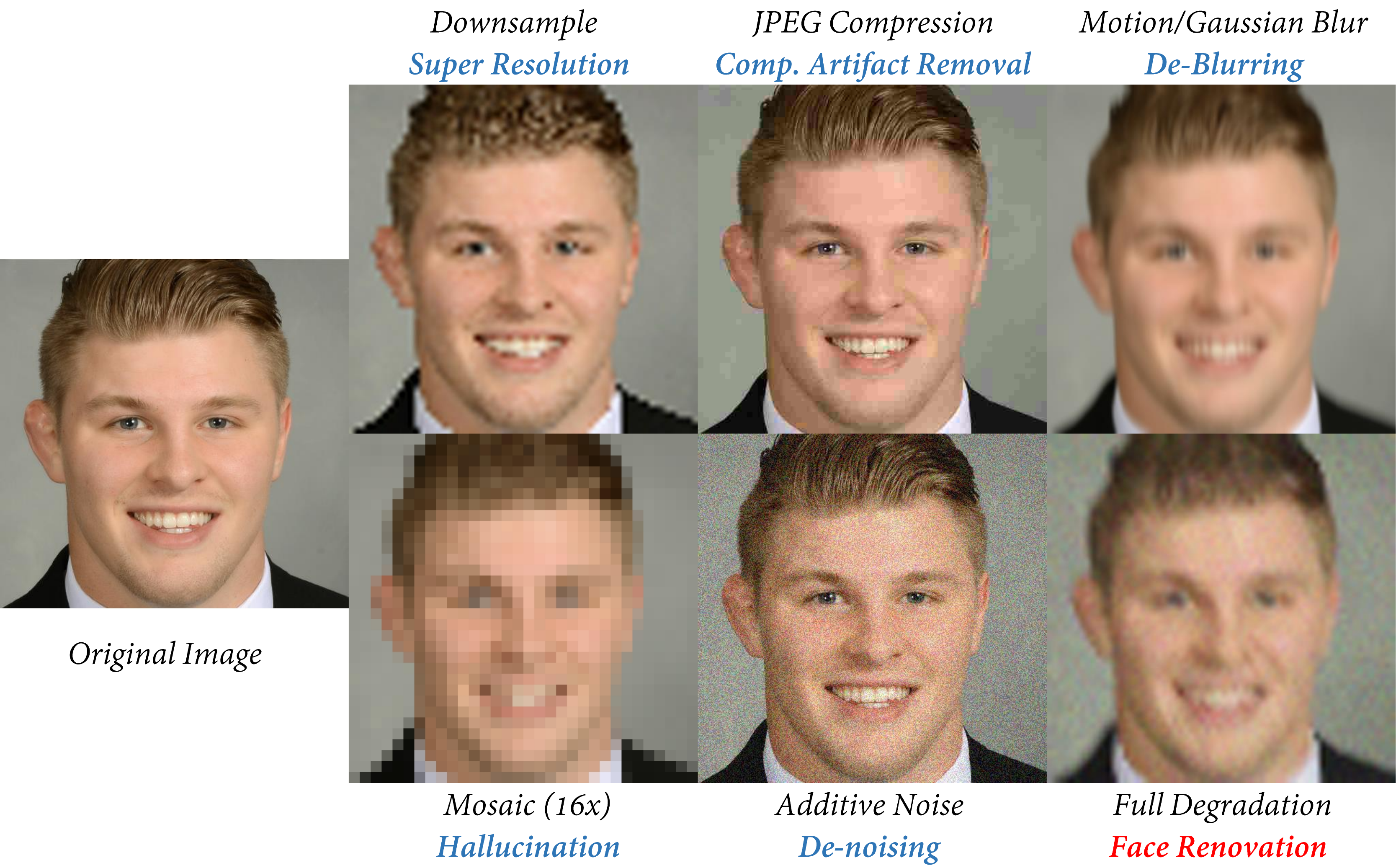}
	\caption{Visualization of degradation types and the corresponding face manipulation tasks. }
	\label{fig:deg_egs}
\end{figure}

\begin{figure*}[!pt]
	\centering
	\includegraphics[width=0.9\linewidth]{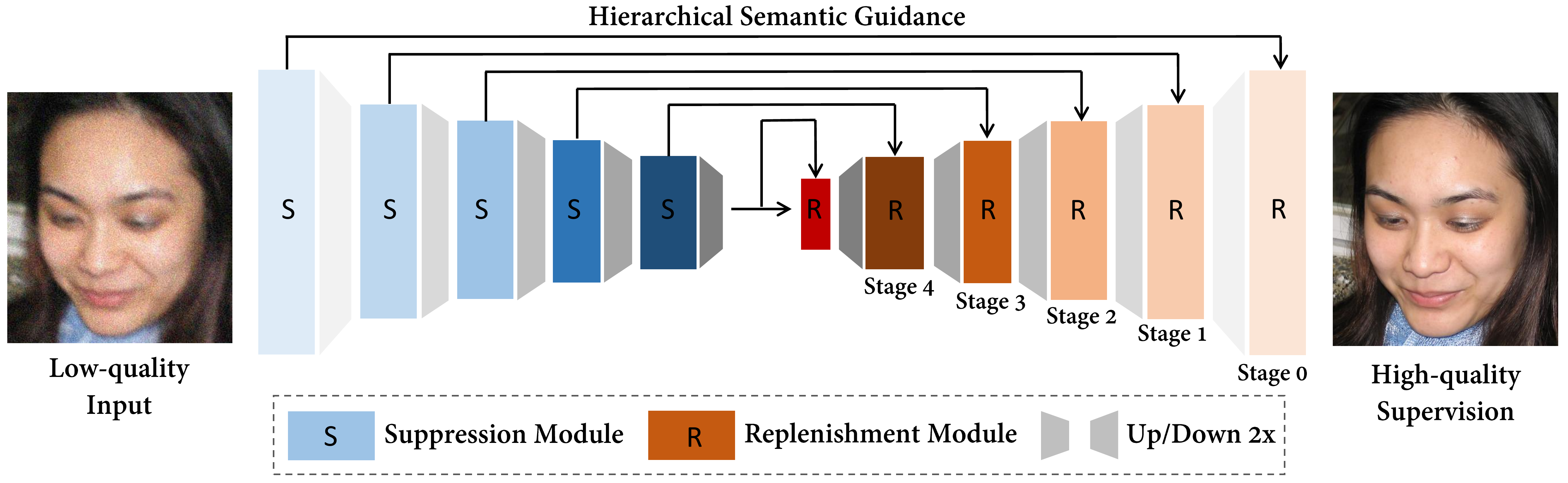}
	\caption{The nested multi-stage architecture of the proposed HiFaceGAN.}\label{fig:network}
\end{figure*}

\subsection{Evaluation Criterion}\label{sec:eval}
For real-world applications, the evaluation criterion for face renovation should be more consistent with human perception rather than machine judgement. Therefore, besides commonly-adopted PSNR and SSIM~\cite{SSIM}\cite{MultiscaleSSIM} metrics, the evaluation criterion for FR should also reflect the semantic fidelity and perceptual realism of renovated faces. For semantic fidelity, we measure the feature embedding distance (FED) and landmark localization error (LLE) with a pretrained face recognition model~\cite{dlib_paper}, where the average L2 norm between feature embeddings is adopted for both metrics. For perceptual realism, we introduce FID~\cite{FID} and LPIPS~\cite{LPIPS} to evaluate the distributional and elementwise distance between original and generated samples in the respective perceptual spaces: For FID it is defined by a pre-trained Inception V3 model~\cite{inceptionv3}, and for LPIPS, an AlexNet~\cite{AlexNet}. Also, the NIQE~\cite{niqe} metric adopted for the 2018 PIRM-SR challenge~\cite{esrgan} is introduced to measure the naturalness of renovated results for in-the-wild face images. Moreover, we will explain the trade-off between statistical and perceptual scores with ablation study detailed in Sec.~\ref{sec:ablation}.

\section{The Proposed HiFaceGAN}

In this section, we detail the architectural design and working mechanism of the proposed HiFaceGAN. As shown in Fig.~\ref{fig:network}, the suppression modules aim to suppress heterogeneous degradation and encode robust hierarchical semantic information to guide the subsequent replenishment module to reconstruct the renovated face with corresponding photorealistic details. Further, we will illustrate the multi-stage renovation procedure and the functionality of individual units in Fig.~\ref{fig:interp} to justify the proposed methodology and provide new insights to the face renovation task.

\subsection{Network Architecture}
\label{sec:architecture}
We propose a nested architecture containing several CSR units that each attend to a specific semantic aspect. Concretely, we cascade the front-end suppression modules to extract layered semantic features, in an attempt to capture the semantic hierarchy of the input image. Accordingly, the corresponding multi-stage renovation pipeline is implemented via several cascaded replenishment modules that each attend to the incoming layer of semantics.
Note that the resulted renovation mechanism differs from the commonly-perceived coarse-to-fine strategy as in progressive GAN \cite{progressivegan}\cite{progressive-face-sr}. Instead, we allow the proposed framework to automatically learn a reasonable semantic decomposition and the corresponding face renovation procedure in a completely data-driven manner, maximizing the collaborative effect between the suppression and replenishment modules. More evidence will be provided in Sec.~\ref{sec:understanding}.

\begin{figure}[!tp]
	\centering
	\includegraphics[width=0.9\linewidth]{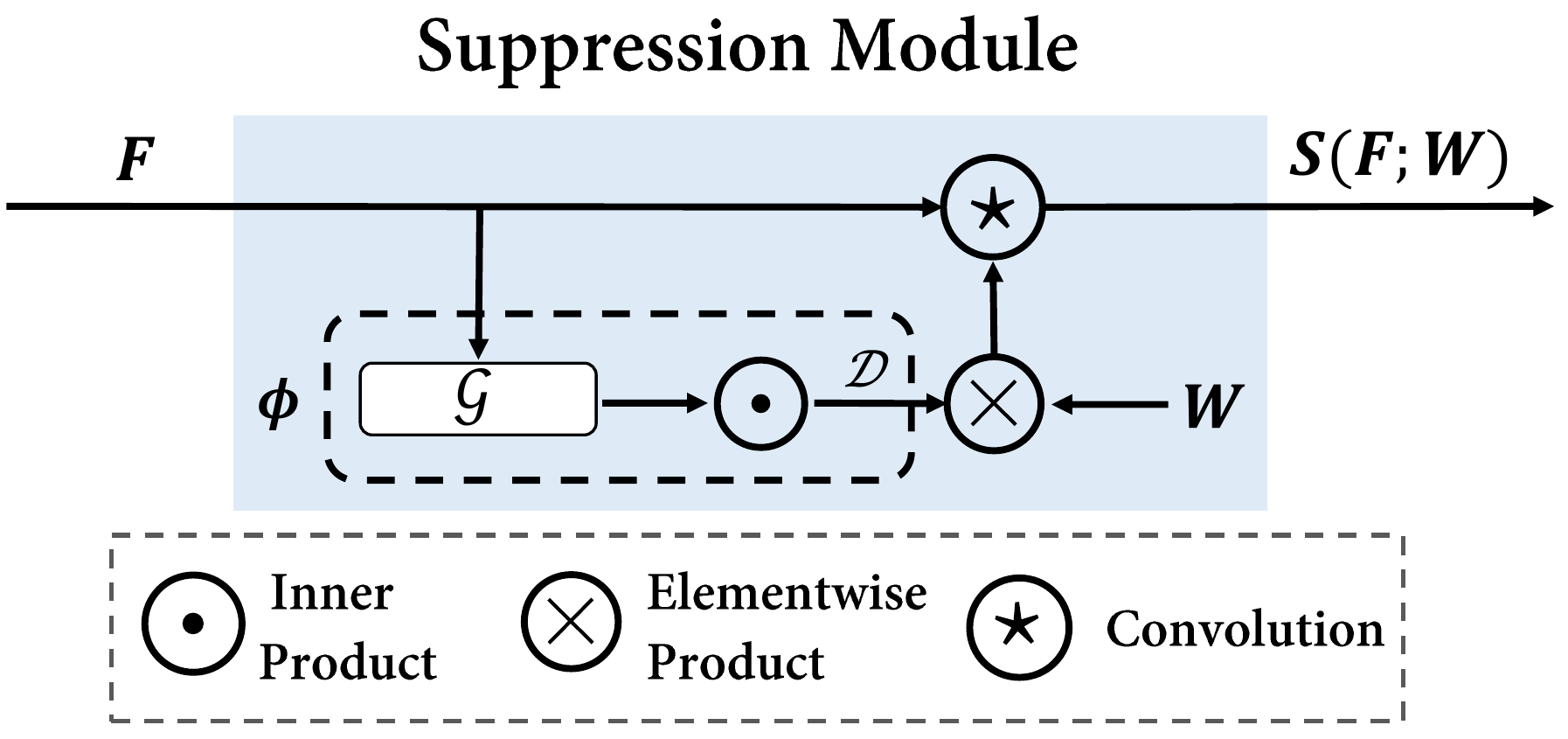}
	\caption{Implementation of the suppression module.}\label{fig:draft}
\end{figure}

\textbf{Suppression Module}
A key challenge for face renovation lies in the heterogeneous degradation mingled within real-world images, where a conventional CNN layer with fixed kernel weights could suffer from the limited ability to discriminate between image contents and degradation artifacts. Let's first look at a conventional spatial convolution with kernel $W\in\mathbb{R}^{C\times C'\times S\times S}$:

\begin{equation}\label{eqn:conv}
\mathrm{conv}(F;W)_{i} = (F * W)_{i} = \sum_{j\in\Omega(i)} w_{\Delta ji}f_j
\end{equation}

where $i,j$ are 2D spatial coordinates, $\Omega(i)$ is the sliding window centering at $i$, $\Delta ji$ is the offset between $j$ and $i$ that is used for indexing elements in $W$. The key observation from Eqn.~\eqref{eqn:conv} is that the conventional CNN layer shares the same kernel weights over the entire image, making the feature extraction pipeline \emph{content-agnostic}. In other words, both the image content and degradation artifacts will be treated in an equal manner and aggregated into the final feature representation, with potentially negative impacts to the renovated image. Therefore, it is highly desirable to select and aggregate informative features with \emph{content-adaptive} filters, such as LIP~\cite{LIP} or PAC~\cite{PixelAdaptiveCNN}. In this work, we implement the suppression module as shown in Fig.~\ref{fig:draft} to replace the conventional convolution operation in Eqn.~\eqref{eqn:conv}, which helps select informative feature responses and filter out degradation artifacts through adaptive kernels. Mathematically,

\begin{equation}\label{eqn:draft}
S(F;W)_{i} = \sum_{j\in\Omega(i)} \phi(f_j,f_i) w_{\Delta ji} f_j
\end{equation}

where $\phi(\cdot,\cdot)$ aims to modulate the weight of convolution kernels with respect to the correlations between neighborhood features. Intuitively, one would expect a correlation metric to be symmetric, i.e. $\phi(f_i,f_j) = \phi(f_j,f_i),\forall f_i, f_j \in \mathbb{R}^{C}$, which can be fulfilled via the following parameterized inner-product function:

\begin{equation}\label{eqn:phi}
\phi(f_i,f_j) = \mathcal{D}(\mathcal{G}(f_i)^\top 
\mathcal{G}(f_j) )
\end{equation}

where $\mathcal{G}: \mathbb{R}^C \rightarrow \mathbb{R}^D$ carries the raw input feature vector $f_i \in \mathbb{R}^C$ into the D-dimensional correlation space to reduce the redundancy of raw input features between channels, and $\mathcal{D}$ is a non-linear activation layer to adjust the range of the output, such as sigmoid or tanh. In practice, we implement $\mathcal{G}$ with a small multi-layer perceptron to learn the modulating criterion in an end-to-end fashion, maximizing the discriminative power of semantic feature selection for subsequent detail replenishment.

\textbf{Replenishment Module}
Having acquired semantic features from the front-end suppression module, we now focus on utilizing the encoded features for guided detail replenishment. Existing works on semantic-guided generation have achieved remarkable progress with spatial adaptive denormalization (SPADE)~\cite{spade}, where semantic parsing maps are utilized to guide the generation of details that belong to different semantic categories, such as sky, sea, or trees. We leverage such progress by incorporating the SPADE block into our cascaded CSR units, allowing effective utilization of encoded semantic features to guide the generation of fine-grained details in a hierarchical fashion. In particular, the progressive generator contains several cascaded SPADE blocks, where each block receives the output from the previous block and replenish new details following the guidance of the corresponding semantic features encoded with the suppression module. In this way, our framework can automatically capture the global structure and progressively filling in finer visual details at proper locations even without the guidance of additional face parsing information.

\subsection{Loss Functions}\label{sec:loss}
Most face restoration works aims to optimize the mean-square-error (MSE) against target images~\cite{srcnn}\cite{vdsr}\cite{edsr}, which often leads to blurry outputs with insufficient amount of details~\cite{esrgan}.
Corresponding to the evaluation criterion in Sec.~\ref{sec:eval}, it is crucial that the renovated image exhibits high semantic fidelity and visual realism, while slight signal-level discrepancies are often tolerable. To this end, we follow the adversarial training scheme~\cite{gan} with an adversarial loss $\L_{GAN}$ to encourage the realism of renovated faces. Here we adopt the LSGAN variant~\cite{lsgan} for better training dynamics:
\begin{equation}\label{eqn:GAN}
\L_{GAN} = \mathrm{E}[\|\log(D(I_{gt}) - 1\|_2^2] + \mathrm{E}[\|\log(D(I_{gen})\|_2^2]
\end{equation}

Also, we introduce the multi-scale feature matching loss $\L_{FM}$~\cite{pix2pixHD} and the perceptual loss $\L_{perc}$\cite{perceptual_loss} to enhance the quality and visual realism of facial details:
\begin{equation}\label{eqn:fm}
\L(\phi) = \sum_{i=1}^{L} \frac{1}{H_i W_i C_i} \|\phi_i(I_{gt}) - \phi_i(I_{gen})\|_2^2
\end{equation}

where for the adversarial loss $\L_{FM}$, $\phi$ is implemented via the multi-scale discriminator $D$ in~\cite{pix2pixHD} and for the perceptual loss $\L_{perc}$, a pretrained VGG-19~\cite{vgg} network.
Finally, combining Eqn.~\eqref{eqn:GAN} and~\eqref{eqn:fm} leads to the final training objective:
\begin{equation}\label{eqn:recon}
\L_{recon} = \L_{GAN} + \lambda_{FM}\L_{FM} + \lambda_{perc}\L_{perc}
\end{equation}

\begin{figure}[!t]
	\centering
	\includegraphics[width=\linewidth]{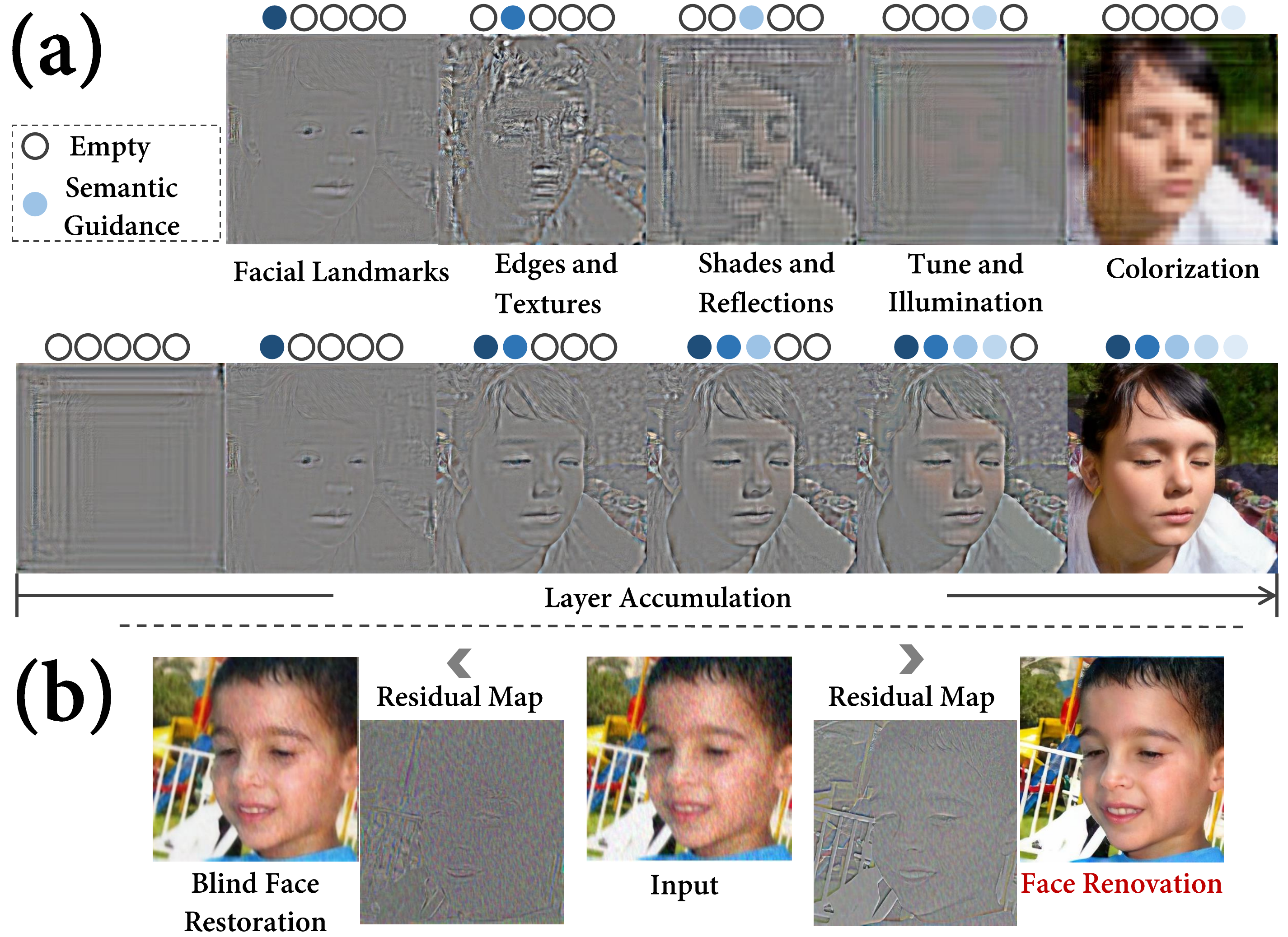}
	\caption{Visualization of (a) the working mechanism of HiFaceGAN and (b) its advantages against existing face restoration works.
	}\label{fig:interp}
\end{figure}



\subsection{Discussion}\label{sec:understanding}
\textbf{The Working Mechanism of HiFaceGAN} To illustrate what each CSR unit can generate at the corresponding stage and how they work cooperatively to perform face renovation outstandingly, we provide an illustrative example shown in Fig.~\ref{fig:interp}(a), where we ablate certain units by replacing the corresponding semantic feature map (blue dots) with a constant tensor (hollow circles), leading to a plain grey background to better isolate the contents generated at each individual stage. Given a 16x down-sampled low-quality facial image, we first sequentially utilize single semantic guidance from the inner stage to the outer stage, the upper row in Fig.~\ref{fig:interp}, and then show the results of the accumulation of semantic guidance in the lower row of Fig. ~\ref{fig:interp}. It is impressive that single semantic guidance from a specific stage leads the corresponding replenishment module to generate a hierarchical layer, which from the inner stage to the outer stage focuses on facial landmarks, edges and textures, shades and reflections, tune and illumination, colorization respectively. In details, by progressively adding semantic guidance, it can be found that with larger receptive field and high-level semantic features, our HiFaceGAN sketches the rough face boundary and localizes facial landmarks, allowing the subsequent CSR unit to replenish fine details upon the basic facial structure when the receptive field size goes down and the resolution raises up. The step-by-step face renovation process acts like a hierarchical layer-by-layer overlaying of contents generated with replenishment modules in a semantic-guided fashion, which gradually enhances the visual quality and realism of the renovated image. So far, the progressive face renovation process with logically reasonable ordered steps has justified our heuristics in the network architecture design and illustrate the efficacy and interpretability of HiFaceGAN in a convincing manner.

\begin{figure*}[!tp]
\centering
\includegraphics[width=\linewidth]{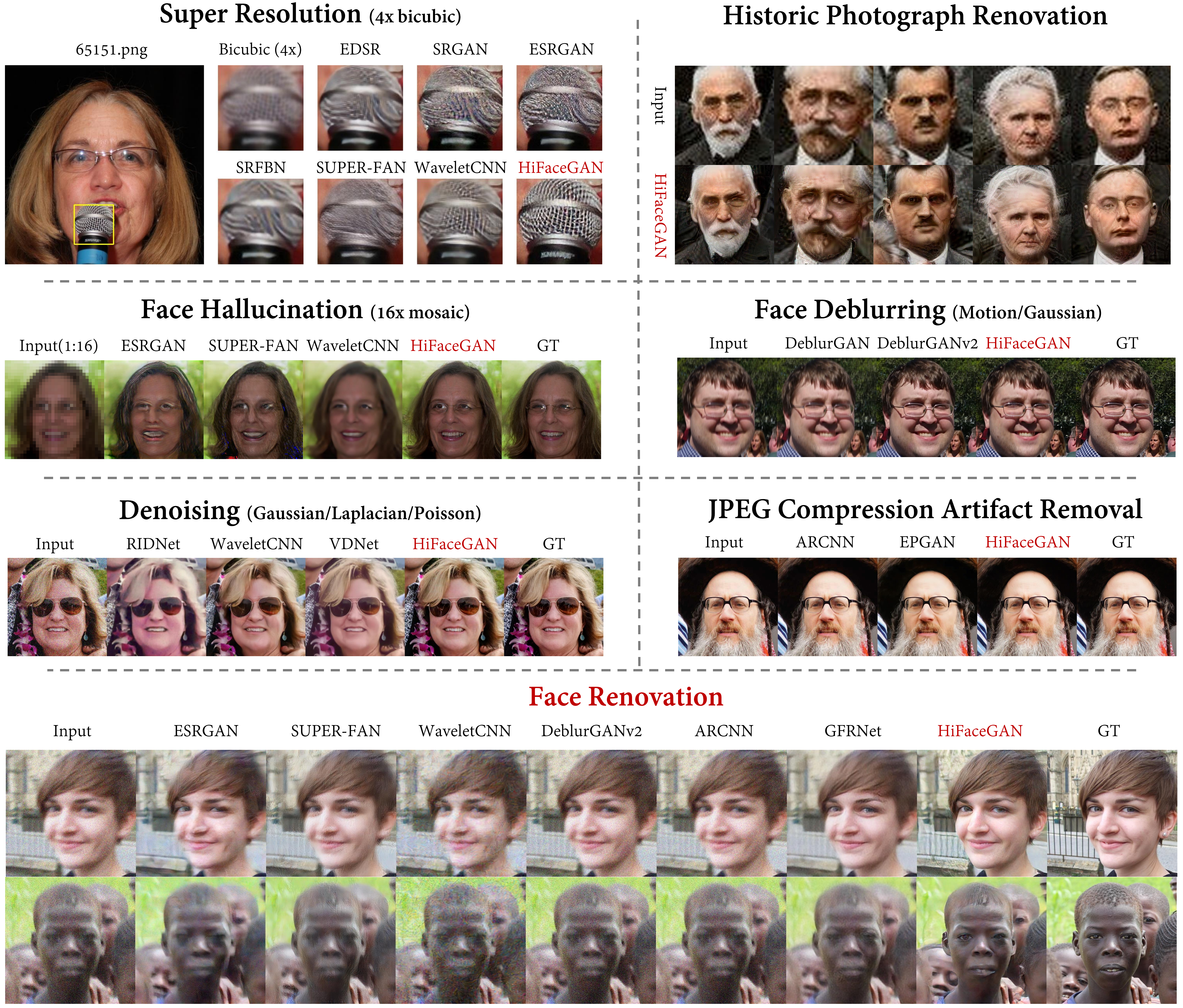}
\caption{
	Qualitative results of our HiFaceGAN and related state-of-the-art methods on all mentioned sub-tasks. \emph{(Best to view on the computer screen for your convenience to zoom in and compare the quality of visual details.)}
}\label{fig:big_figure}
\end{figure*}

\textbf{Comparison with Blind Face Restoration} To better clarify the distinctions between our face renovation framework and existing restoration methods, we compare the residual maps generated with our HiFaceGAN and the state-of-the-art blind face restoration network GFRNet~\cite{BlindFR-ECCV2018}. As shown in Fig.~\ref{fig:big_figure}(b), the residual map generated with GFRNet packs heavier noise and less semantically meaningful details, indicating a higher focus on ``suppression'' and insufficient attention to ``replenishment''. This could be attributed to the PSNR-oriented optimization objective, where additive noises contribute a large proportion of the signal discrepancy. In contrast, HiFaceGAN can simultaneously suppress degradation artifacts and replenish semantic details, leading to semantic-aware residual maps and more refined renovation results. Also, HiFaceGAN can renovate background contents and foreground faces together, leading to consistent quality improvement across the entire image. This justifies the rationale of the ``dual-blind'' setting towards real-world applications with images containing rich non-facial contents.


\begin{table*}[!pt]
	\centering
	\caption{Quantitative comparisons to the state-of-the-art methods on the newly proposed face renovation and five related face manipulation tasks.
		\emph{(Up arrow means the higher score is preferred, and vice versa.)}}
	\label{tab:compare_sota_sr}
	\resizebox{0.85\textwidth}{!}{
		\begin{tabular}{|c|c|ccc|cc|ccc|}
			\hline
			& & & Statistical & &\multicolumn{2}{c|}{Semantic} & \multicolumn{3}{c|}{Perceptual}\\
			Task & Methods & PSNR $ \uparrow $ &  SSIM $ \uparrow $ & MS-SSIM $ \uparrow $ & FED $ \downarrow $ & LLE $ \downarrow $ & FID $ \downarrow $ & LPIPS $ \downarrow $ & NIQE $ \downarrow $ \\
			\hline
			& EDSR~\cite{edsr} & 30.188 & 0.824 & 0.961 & 0.0843 & \textbf{2.003} & 20.605 & 0.2475 & 13.636 \\
			& SRGAN~\cite{srgan} & 27.494 & 0.735 & 0.935 & 0.1097 & 2.269 & 4.396 & 0.1313 & 7.378 \\
			\textbf{Face Super} & ESRGAN~\cite{esrgan} & 27.134 & 0.741 & 0.935 & 0.1107 & 2.261 & 3.503 & 0.1221 & 6.984 \\
			\textbf{Resolution}  & SRFBN~\cite{feedback_sr} & 29.577 & 0.827 & 0.953 & 0.0984 & 2.066& 20.032 & 0.2406 & 13.901 \\
			(4x, Bicubic)& Super-FAN~\cite{SuperFAN} & 25.463 & 0.729 & 0.913 & 0.1416 & 2.333 & 14.811 & 0.2357 & 8.719 \\
			& WaveletCNN~\cite{waveletcnn_conf} & 28.750 & 0.806 & 0.952 & 0.0964 & 2.072 & 16.472 & 0.2443 & 12.217 \\
			& \textbf{HiFaceGAN} & \textbf{30.824} & \textbf{0.838} & \textbf{0.971} & \textbf{0.0716} & 2.071& \textbf{1.898} & \textbf{0.0723} & \textbf{6.961} \\
			\hline\hline
			
			& Super-FAN & 20.536 & 0.540 & 0.744 & 0.4297 & 4.834& 63.693 & 0.4411 & 7.444 \\
			\textbf{Hallucination} & ESRGAN & 21.001 & 0.576 & 0.697 & 0.5138 & 5.902 & 50.901 & 0.3928 & 15.383 \\
			(16x, Mosaic) &WaveletCNN & \textbf{23.810} & \textbf{0.675} & \textbf{0.837} & 0.3713 & 3.729 & 60.916 & 0.4909 & 11.450 \\
			& \textbf{HiFaceGAN} & 23.705 & 0.619 & 0.819 & \textbf{0.3182} & \textbf{3.137} & \textbf{11.389} & \textbf{0.2449} & \textbf{6.767} \\
			\hline\hline
			
			\textbf{Denoising} & RIDNet~\cite{RIDNet} & 25.432 & 0.731 & 0.891 & 0.2128 & 2.465& 36.515 & 0.3864 & 13.002 \\
			(1/3 Gaussian, & WaveletCNN & 26.530 & 0.754 & 0.895 & 0.2441 & 2.728 & 26.731 & 0.3119 & 11.395 \\
			1/3 Poisson, & VDNet~\cite{VDNet} & 27.718 & 0.797 & 0.928 & 0.1551 & 2.297 & 15.826 & 0.2458 & 14.262 \\
			1/3 Laplacian) & \textbf{HiFaceGAN} & \textbf{31.828} & \textbf{0.845} & \textbf{0.957} & \textbf{0.1109} & \textbf{2.090} & \textbf{3.926} & \textbf{0.0868} & \textbf{7.341} \\
			\hline\hline
			
			\textbf{Deblurring} & DeblurGAN~\cite{DeblurGAN} & 25.304 & 0.718 & 0.894 & 0.1786 & 3.219 & 14.331 & 0.2574 & 12.697 \\
			(1/2 Motion blur& DeblurGANv2~\cite{DeblurGANv2} & 26.908 & 0.773 & 0.913 & 0.1043 & 3.036 & 10.285 & 0.2178 & 13.729 \\
			1/2 Gaussian blur)& \textbf{HiFaceGAN} & \textbf{28.928} & \textbf{0.793} & \textbf{0.954} & \textbf{0.0913} & \textbf{2.156} & \textbf{2.580} & \textbf{0.0874} & \textbf{7.426} \\
			\hline\hline
			
			& ARCNN~\cite{ARCNN} & \textbf{33.021} & 0.879 & 0.972 & 0.0845 & \textbf{1.959} & 9.761 & 0.1551 & 14.827 \\
			\textbf{JPEG artifact} & EPGAN~\cite{EPGAN} & 32.780 & \textbf{0.882} & \textbf{0.976} & \textbf{0.0814} & 1.979 & 10.250 & 0.1638 & 13.729 \\
			\textbf{removal} & \textbf{HiFaceGAN} & 31.611 & 0.850 & 0.970 & 0.0842 & 2.057& \textbf{1.880} & \textbf{0.0541} & \textbf{6.911} \\
			\hline\hline
			
			& Degraded Input & 22.905 & 0.465 & 0.756 & 0.2875 & 3.936 & 63.670 & 0.6828 & 21.955 \\
			& Super-FAN & 24.818 & 0.549 & 0.818 & 0.2495 & 3.705 & 32.800 & 0.4283 & 12.154 \\
			& ESRGAN & 24.197 & 0.564 & 0.816 & 0.2761 & 3.771 & 28.053 & 0.4141 & 11.382 \\
			\textbf{Face Renovation} & WaveletCNN & 24.404 & 0.648 & 0.817 & 0.2821 & 3.690 & 58.901 & 0.3102 & 15.530 \\
			(Full Degradation) & DeblurGANv2 & 23.704 & 0.494 & 0.776 & 0.2403 & 4.412 & 49.329 & 0.6496 & 21.983 \\
			& ARCNN & 24.187 & 0.539 & 0.787 & 0.2580 & 3.833 & 60.864 & 0.6424 & 18.880 \\
			& GFRNet~\cite{BlindFR-ECCV2018} & 25.227 & \textbf{0.686} & 0.854 & 0.2524 & 3.371 & 48.229 & 0.4591 & 20.777 \\
			& \textbf{HiFaceGAN} & \textbf{25.837} & 0.674 & \textbf{0.881} & \textbf{0.2055} & \textbf{2.701} & \textbf{8.013} & \textbf{0.2093} &\textbf{7.272} \\
			\hline\hline
			
			\textbf{Real Image} & --- & $+\infty$ & 1 & 1 & 0 & 0& 0 & 0 & 7.796 \\
			\hline
			
		\end{tabular}
	}
\end{table*}

\begin{table}[!t]
	\centering
	\caption{Ablation study results on 16x face hallucination.}
	\label{tab:ablation}
	\resizebox{0.45\textwidth}{!}{
		\begin{tabular}{|c|ccccc|}
			\hline
			Metrics & SPADE & 16xFace & FixConv & Default & L1\\
			\hline
			PSNR $ \uparrow $ & 20.968 & 23.541 & 23.651 & 23.705 & \textbf{23.937} \\
			SSIM $ \uparrow $ & 0.596 & 0.610 & 0.615 & 0.619 & \textbf{0.628}\\
			MS-SSIM $ \uparrow $ & 0.718 & 0.811 & 0.818 & 0.819 & \textbf{0.823} \\
			\hline
			FED $ \downarrow $ & 0.4595 & 0.3296 & 0.3236 & \textbf{0.3182} & 0.3197\\
			LLE $ \downarrow $ & 4.143 & 3.227 & 3.157 & 3.137 & \textbf{3.085} \\
			\hline
			FID $ \downarrow $ & 52.701 & 14.365 & 13.154 & \textbf{11.389} & 11.910\\
			LPIPS $ \downarrow $ & 0.3967 & 0.2609 & 0.2467 & \textbf{0.2449} & 0.2462\\
			NIQE $ \downarrow $ & 10.367 & 7.446  & 7.011 & \textbf{6.767} & 6.938 \\
			
			\hline
		\end{tabular}
	}
\end{table}

\section{Experiments}\label{sec:experiments}
In this section, we demonstrate the versatility, robustness and generalization ability of our proposed HiFaceGAN over a wide range of related face restoration sub-tasks, both on synthetic images and real-world photographs. Furthermore, we conduct an ablation study to verify our major contributions and stimulate future research directions. Detailed configurations are provided in supplementary materials to facilitate the reproduction.

\subsection{Comparison with state-of-the-art methods}\label{sec:sota}
We first evaluate our framework on five subtasks, including super resolution, hallucination, denoising, deblurring and compression artifact removal. For each subtask, the training data is prepared by performing task-specific degradation upon raw images from FFHQ~\cite{stylegan_ffhq}, Fig.~\ref{fig:deg_egs}\footnote{All baselines are retrained using default settings unless training code is unavailable.}. Finally, five most competitive task-specific methods, along with the state-of-the-art blind face restoration baseline~\cite{BlindFR-ECCV2018}, are chosen to compete with our HiFaceGAN over the most challenging and practical FR task.

\textbf{Comparison with Task-Specific Baselines}
Overall, HiFaceGAN outperforms all baselines by a huge margin on perceptual performance, with 3-10 times gain on FID and 50\%-200\% gain on LPIPS, Table~\ref{tab:compare_sota_sr}. Furthermore, HiFaceGAN even outperform real images in terms of naturalness, as reflected by the NIQE metric. Generally, our generative approach is better suited for tasks with heavy structural degradation, such as face hallucination, denoising and deblurring. For super-resolution and JPEG artifact removal, the structural degradation is considerably milder, Fig.~\ref{fig:deg_egs}, leading to narrowed gaps between task-specific solutions and our generalized framework, especially on statistical scores. This is reasonable since the training functions are more perceptually inclined for FR. Nevertheless, it is still possible to trade-off between perceptual and statistical performance, as will be discussed in ablation study.

For qualitative comparison, we showcase the representative results on corresponding tasks in Fig.~\ref{fig:big_figure}. For all subtasks, our HiFaceGAN can replenish rich and convincing visual details, such as hair bangs, beards and wrinkles, leading to consistent, photo-realistic renovation results. In contrast, other task-specific methods either produce over-smoothed or color-shifted results (WaveletCNN, SIDNet), or incur severe systematic artifacts during detail replenishment (ESRGAN, Super-FAN). Moreover, our dual-blind setting is equally effective in enhancing details for non-facial contents, such as the interweaving grids on the microphone. In summary, HiFaceGAN can resolve all types of degradation in a unified manner with stunning renovation performances, verifying the efficacy of the proposed methodology and architectural design. More results are provided in the supplementary material.

\textbf{Dual-Blind vs. Single-Blind}
To discuss the impact of external guidance, we compare our HiFaceGAN with the ``single-blind'' GFRNet~\cite{BlindFR-ECCV2018} over the fully-degraded FFHQ datset, where the ground truth image is provided as the high-quality guidance during testing. As shown in column 7-9 of Fig.~\ref{fig:big_figure}, even with such a strong guidance, GFRNet is still less effective in suppressing noises and replenishing fine-grained details than our network, indicating its limitation in feature utilization and generative capability. Consistent with our observation in Sec.~\ref{sec:understanding}, the performance gain of GFRNet against other baselines is mainly statistical, where the semantic and perceptual scores are less competitive, Table~\ref{tab:compare_sota_sr}. Our empirical study suggests that 1) the lack of explicit guidance does not necessarily lead to inferior performance of face renovation; 2) the ability to replenish plausible details is most crucial for high-quality face renovation.

\subsection{Historic Photograph Renovation}\label{sec:HPR}

The historic group photograph of famous physicists at the contemporary age taken at the 5th Solvay Conference in 1927 is utilized to evaluate generalization capability of state-of-the-art models for real-world face renovation, Fig.~\ref{fig:teaser}. We crop $64\times64$ face patches from the original image and resize them to $512\times512$ with bicubic interpolation for input. Apparently, compared to others, our HiFaceGAN can successfully suppress complex degradation in real old photos to generate faces with high definition, high fidelity, and fewer artifacts, while replenishing realistic details, such as facial luster, fine hair, clear facial features, and photo-realistic wrinkles. More outstanding renovation results are displayed in Fig.~\ref{fig:big_figure}. Inevitably, the renovated faces contain minor artifacts that mostly occur at shading regions, where degradation artifacts have severely corrupted the underlying contents. Nevertheless, the renovated high-resolution person portraits still possess much better visual and artistic quality than the original input, which simultaneously demonstrates the capability of our model and the challenge in real-world applications.

\subsection{Ablation Study}\label{sec:ablation}

We perform an ablation study over the most challenging 16x face hallucination task to verify three aspects of our framework: guidance type, architecture, and component design. The four ablation methods are described below:
\begin{itemize}
	\item \textbf{SPADE} The vanilla SPADE~\cite{spade} network with semantic guidance being face parsing maps extracted from the original \textit{high-resolution} images with a pretrained parsing model~\cite{MaskGAN}.
	\item \textbf{16xFace} replaces the semantic parsing map in SPADE with degraded faces containing 16-pixel mosaics.
	\item \textbf{FixConv} retains the nested CSR architectue of HiFaceGAN with the normal content-agnostic convolution layer in Eqn~\eqref{eqn:conv}.
	\item \textbf{L1} adds an additional L1 loss upon default HiFaceGAN to adjust between statistical and perceptual scores.
\end{itemize}

The evaluation scores are reported in Table~\ref{tab:ablation}. Although face parsing maps provide much finer spatial guidance, it is evident that face renovation relies more on semantic features, as reflected by the huge performance gap between SPADE and 16xFace. Also, FixConv achieves visible performance gain by extracting hierarchical semantic features and applying multi-stage face renovation, verifying the proposed nested architecture. Moreover, incorporating the content-adaptive suppression module further improves the feature selection and degradation suppression ability, leading to substantial gain over FixConv on perceptual and semantic scores. Finally, adding an L1 loss term makes the model statistically inclined, with superior PSNR/SSIM and inferior FID/LPIPS/NIQE scores, verifying the flexibility of our framework to trading off between statistical and perceptual performances.

\section{Conclusion and Future Work}
In this paper, we present a challenging, yet more practical task towards real-world photo repairing applications, termed as ``face renovation''. Particularly, we propose HiFaceGAN, a collaborative suppression and replenishment framework that works in a ``dual-blind'' fashion, reducing dependence on degradation prior or structural guidance for training. Extensive experiments on both synthetic face images and real-world historic photographs have demonstrated the versatility, robustness and generalization capability over a wide range of face restoration tasks, outperforming current state-of-the-art by a large margin. Furthermore, the working mechanism of HiFaceGAN, and the rationality of the ``dual-blind'' setting are justified in a convincing manner with illustrative examples, bringing fresh insights to the subject matter. 
In the future, we envision that the proposed HiFaceGAN would serve as a solid stepping stone towards the expectations of face renovation. Specifically, the severe degradation often lead to content ambiguity for renovation, like the Afro haircut appeared in Fig.~\ref{fig:big_figure} where our method misjudged as normal straight hairs, which motivates us to increase the diversity and balance between different ethnic groups during data collection. Also, it is still a huge challenge for the renovation of objects with regular geometric shapes (such as glasses) and partially-occluded faces --- a typical case where external structural guidance could be beneficial. Therefore, exploring multi-modal generation networks with both structural and semantical guidance is another possibility.

\begin{acks}
	This work is done during Lingbo's internship at DAMO Academy, Alibaba Group, and supported by the National Natural Science Foundation of China (61931014, 61632001) and High-performance Computing Platform of Peking University, which are gratefully acknowledged.
\end{acks}

\bibliographystyle{ACM-Reference-Format}
\bibliography{sample-base}

\end{document}